\documentclass[11pt]{article}

\usepackage[preprint]{acl}

\usepackage{times}
\usepackage{latexsym}
\usepackage{url}
\usepackage[T1]{fontenc}

\usepackage[utf8]{inputenc}

\usepackage{microtype}

\usepackage{inconsolata}

\usepackage{graphicx}

\title{Efficient Context-Limited Telescope Bibliography Classification for the WASP-2025 Shared Task Using SciBERT\thanks{Our code is available at \url{https://github.com/E0NIA/TRACS-WASP-2025-1st-Place}.}}

\author{Madhusudhana Naidu \\
\texttt{mnaidu1025@gmail.com}} 

\begin{document}
\maketitle
\begin{abstract}
The creation of telescope bibliographies is a crucial part of assessing the scientific impact of observatories and ensuring reproducibility in astronomy. This task involves identifying, categorizing, and linking scientific publications that reference or use specific telescopes. However, this process remains largely manual and resource intensive. In this work, we present an efficient SciBERT-based approach for automatic classification of scientific papers into four categories — science, instrumentation, mention, and not telescope. Despite strict context-length constraints (maximum 512 tokens) and limited compute resources, our approach achieved a macro F1 score of 0.89, ranking at the top of the WASP-2025 leaderboard. We analyze the effect of truncation and show that even with half the samples exceeding the token limit, SciBERT’s domain alignment enables robust classification. We discuss trade-offs between truncation, chunking, and long-context models, providing insights into the efficiency frontier for scientific text curation.
\end{abstract}
\noindent 
\textbf{Keywords:} Scientific Document Processing, Multi-label Classification, SciBERT, Bibliography Curation, Astronomy, Context Limitation.

\section{Introduction}

The assessment of the scientific impact of observational facilities often relies on bibliometric analyses of research publications that use data from those telescopes. Creating and maintaining these bibliographies requires identifying relevant papers, disambiguating telescope mentions, and classifying the nature of data use — a process still largely performed manually. Automating this process would significantly benefit librarians, archivists, and research scientists by improving reproducibility and discoverability of astronomical data.

The WASP-2025 Shared Task \citep{grezes2025kaggle, grezes-etal-2025-tracs} aims to develop AI assistants capable of automating this bibliography curation. Given textual data from scientific papers—including title, abstract, body, acknowledgments, and grants—participants were asked to identify the telescope referenced and classify each paper as science, instrumentation, mention, or not telescope.

Large language models (LLMs) are capable of understanding complex scientific semantics, but applying them efficiently under strict computational and input-length constraints remains challenging. In this work, we focus on designing a lightweight yet effective SciBERT-based model \citep{beltagy2019scibert} that can operate within a 512-token window, significantly below the combined 100k token context of the combined row sample.

Our contributions:
\begin{itemize}
    \item We demonstrate that domain-specific pretraining (SciBERT) can outperform large-scale general models in constrained settings.
    \item We empirically analyze the trade-off between token truncation and classification performance.
    \item We achieve top leaderboard performance (F1 = 0.89) using only Kaggle GPU resources.
\end{itemize}
\section{Task and Dataset}

The task consists of identifying whether a paper refers to a telescope and classifying its relationship to that telescope into one or more of four labels: science, instrumentation, mention, and not telescope.
Each record in the dataset includes:
\begin{itemize}
    \item \textbf{Textual fields:} title, abstract, body, acknowledgments, and grants.
    \item \textbf{Metadata:} author, year, and a unique bibcode.
    \item \textbf{Target labels:} science, instrumentation, mention, and not telescope, requiring multi-label classification.
\end{itemize}
\begin{table*}[t]
\centering
\caption{Token length statistics for key textual fields excluding body using the SciBERT tokenizer. The median (50\%), 75th, and 95th percentiles are shown, highlighting that the abstract and acknowledgments sections often exceed typical model input limits.}
\label{tab:token_stats}
\begin{tabular}{lrrrr}
\hline
\textbf{Field} & \textbf{Mean} & \textbf{Median (50\%)} & \textbf{75th Percentile} & \textbf{95th Percentile} \\ \hline
Title           & 18.3          & 17.0                   & 22.0                     & 32.0                     \\
Abstract        & 337.6         & 325.0                  & 419.0                    & 631.0                    \\
Acknowledgments & 163.8         & 106.0                  & 228.0                    & 554.0                    \\
Grants          & 0.8           & 0.0                    & 0.0                      & 7.0                      \\ \hline
\end{tabular}
\end{table*}
The training data exhibits a significant class imbalance. The frequencies for each positive label are as follows:
\begin{itemize}
    \item \texttt{science}: 37,881
    \item \texttt{mention}: 34,813
    \item \texttt{not\_telescope}: 7,772
    \item \texttt{instrumentation}: 875
\end{itemize}


A primary challenge of this task is the extensive length of the input text. As quantified in Table~\ref{tab:token_stats}, a substantial number of samples contain text sections that individually exceed the 512-token context window of standard transformer models. The combined text from all fields when including body can surpass 50,000 tokens, creating a severe context limitation and motivating our approach of using an efficient, truncated-context model.

\section{Methodology}

\subsection{Baseline: TF-IDF + Logistic Regression}
As a baseline, we implemented a traditional machine learning pipeline combining TF-IDF vectorization with a One-vs-Rest Logistic Regression classifier. Each sample was represented using the concatenation of its title, abstract, acknowledgments, and grants sections, separated by [SEP] tokens. The TF-IDF vectorizer was configured with bi-grams (1–2), a vocabulary size of 20,000, and English stopword removal. In addition, one-hot encoding was applied to the telescope categorical feature, and the year was treated as a numeric feature and passed through directly.

The classifier used the liblinear solver with class-balanced weighting to handle label imbalance, and was wrapped in a One-vs-Rest strategy to support multi-label classification across the four categories (science, instrumentation, mention, not telescope). The model was trained on an 80/20 train–validation split. This baseline achieved a macro F1 score of 0.66 on the training set and 0.82 on the test leaderboard, providing a strong benchmark for subsequent transformer-based experiments.

\subsection{SciBERT with Truncated Context}
Our best-performing system was based on SciBERT (allenai/scibert scivocab uncased), fine-tuned for multi-label classification over the four task categories: science, instrumentation, mention, and not telescope. The input text was constructed by concatenating the telescope name, year, title, abstract, acknowledgments, and grants fields using special [SEP] separators. All missing text fields were replaced with empty strings to ensure consistency. Data were split into training and validation sets (80/20), and tokenized using the SciBERT tokenizer with a maximum sequence length of 512 tokens, truncating any longer samples.

The model was trained using AdamW optimizer with a learning rate of 2e-5, batch size 48, and 3 epochs on the Kaggle 2 * T4 GPU (15 GB VRAM). A BCEWithLogitsLoss function was used to accommodate the multi-label nature of the task, and learning rate scheduling was handled via a linear scheduler with no warmup. Training was distributed using DataParallel for multi-GPU availability. The best model was selected based on macro F1 score on the validation set, and checkpointed whenever improvement was observed. Despite truncation of roughly 50\% of samples exceeding 512 tokens, this configuration achieved robust generalization, reaching a leaderboard F1 of 0.89, indicating strong adaptation of domain-specific representations for telescope bibliography classification.

\subsection{Potential Extensions}
Given more time and compute, two extensions could be performed:

Chunked Input Windows: Breaking long documents into overlapping windows (stride = 128) for majority voting or mean pooling of predictions.

Longformer Backbone: Leveraging 4096-token context to capture extended information from the abstract and acknowledgement sections.

\section{Results}

See Table~\ref{tab:model_results} for the results of our model and the baselines.
Even though SciBERT processed less than half of the full context, it outperformed models capable of handling longer inputs.
This suggests that key discriminative signals are concentrated in the title, abstract, and acknowledgments.

The truncation robustness of SciBERT highlights the power of domain-specific pretraining, particularly when resources are limited.

\begin{table*}[t]
\centering
\caption{Macro F1 scores for our model and baselines on the validation set (CV) and the final leaderboard (LB).}
\label{tab:model_results} 
\begin{tabular}{llrr}
\hline
\textbf{Model} & \textbf{Context} & \textbf{F1 (CV)} & \textbf{F1 (LB)} \\ \hline
TF-IDF + OVR     & NA               & 0.66             & 0.82             \\
SciBERT          & 512              & 0.80             & 0.89             \\ \hline
Random Baseline  & NA               & NA               & 0.24             \\
gpt-oss20b       & NA               & NA               & 0.31             \\ \hline
\end{tabular}
\end{table*}
\section{Discussion}
The results demonstrate that domain-specific language models like SciBERT are highly effective for automating telescope bibliography curation, even under significant computational constraints. A key observation is that the acknowledgment section strongly correlates with the presence of telescope-related data, particularly for widely used observatories such as Chandra or Hubble. This suggests that acknowledgment text often encodes implicit evidence of data usage, making it an informative input for classification. However, due to the limited GPU resources available on the Kaggle platform (P100 GPU with 15 GB VRAM and 30 GB CPU RAM), experiments were restricted to a maximum context length of 512 tokens, with longer inputs truncated. Despite this limitation, the model achieved a macro F1 score of 0.89 on the leaderboard, significantly outperforming both the GPT-OSS20B \citep{openai2025modelcard} baseline (0.31) and the random submission (0.24). Longer-context architectures such as Longformer or chunked SciBERT approaches could potentially capture broader contextual signals, especially from the full body text, and further improve classification accuracy.

\section{Conclusion}
This work presents a lightweight yet high-performing approach for classifying telescope-related literature within the WASP-2025 shared task. Starting from a TF-IDF baseline and advancing to a fine-tuned SciBERT model, the system achieved state-of-the-art results while operating within strict computational limits. The findings highlight that concise context—when combined with a domain-trained encoder—can effectively capture scientific intent and data references in astronomy papers. Future extensions may include section-wise modeling, hierarchical encoding, or integration of long-context transformer variants to enhance interpretability and recall. More broadly, this study underscores the potential of AI-assisted systems to support the bibliographic curation and reproducibility efforts of scientific observatories.

\end{document}